\documentclass[a4paper,12pt]{article}

\usepackage{arxiv}

\usepackage[utf8]{inputenc} 
\usepackage[T1]{fontenc}    
\usepackage{hyperref}       
\usepackage{url}            
\usepackage{booktabs}       
\usepackage{amsfonts}       
\usepackage{nicefrac}       
\usepackage{microtype}      
\usepackage{lipsum}

\usepackage{multicol} 
\usepackage[pdftex]{graphicx}

\graphicspath{{noiseimages/}}
\usepackage{enumitem}
\usepackage{amssymb}
\usepackage[nooneline]{caption}

\usepackage{blindtext}
\usepackage{scrextend}
\addtokomafont{labelinglabel}{\sffamily}

\usepackage{amsmath}
\usepackage{enumerate}
\usepackage{multirow}

\usepackage[ruled,vlined]{algorithm2e} 

\usepackage{rotating}

\title{Analysis of basic emotions in texts \\ based on bert vector representation}

\author{
  A.~Artemov,\ A.~Veselovskiy,\ I.~Khasenevich,\ I.~Bolokhov
   \\
 Kognitivnie Sistemi\\
  \texttt{office@cogsys.company} \\
}

\begin{document}
\maketitle


\vspace{1cm}

\begin{abstract}
In the following paper the authors present a GAN-type model and the most important stages of its development for the task of emotion recognition in text. In particular, we propose an approach for generating a synthetic dataset of all possible emotions combinations based on manually labelled incomplete data.
\end{abstract}

\section*{Introduction}
\begin{multicols}{2}
There are several approaches to classification of human emotions. Authors consider the methodology according to P.Ekman to be the most reasonable \cite{4}. In accordance with it, the following seven human emotions can be distinguished: fear, sadness, anger, disgust, calm, happiness, surprise. These emotions were classified according to the analysis of facial expressions of a person during events of different semantic meaning. The unambiguous perception of facial expressions with different people and nations was the proof of the sufficiency and validity of the seven basic emotions.

At the same time, although there is a number of solutions for emotional recognition, a standard for the number of basic emotions for text data has not yet been created. For example, in Rodrigo Masaru Ohashi’s work \cite{7} the model of bidirectional LSTM with a convolutional neural network layer was successfully trained to identify 4 emotions: joy, fear, sadness and anger. Chew-Yean Yam’s article “Emotion Detection and Recognition from Text Using Deep Learning \cite{2}” describes the neural network with 3 hidden layers of 125, 25 and 5 neurons for determining 5 emotions: anger, sadness, fear, happiness, excitement. Jangwon Park \cite{8} created a Huggingface Transformers based model that predicts 7 emotions, the F1-measure of the model is 59.81 and 61.48 for the train and test dataset, respectively. This model is based on the GoEmotions \cite{3} dataset with 27 emotion classes (see Table \ref{tab5}).

Let us assume, as a hypothesis, that the 7 basic emotions by Ekman are applicable to texts. That is, the labelling of texts for this set of emotion classes by a person is complete. Based on the possibility of obtaining such a dataset, the task is to train a model that can predict the vector of seven emotions for an arbitrary text. The result of this work is described in the article.

A positive solution to this problem opens up prospects for combining different data channels (audio and video) to determine emotions when machines are communicating with humans. Moreover, of particular interest is the description of information spread through means of mass communication (social networks), taking into account the phenomenon of emotions social exchange.

\section{Dataset vectorization}

Initially, it is necessary to form a dataset for training the model. The dataset should contain data about seven basic emotions for each certain piece of text - a paragraph or a sentence. The authors couldn’t find a publicly available dataset with this or similar data. Therefore, it was decided to form it automatically using data on the emotional reaction of people, expressed in texts. Such data were emoticons in texts from Twitter, YouTube and other social networks.

The original texts were divided into sentences, emoticons were extracted from each sentence. In order to vectorize emotions, a special dictionary was compiled containing emoticons divided into emotion types according to uniquely interpreted classes:

\end{multicols}

\begin{figure}[h!!]
\centering
\includegraphics[width=0.8\linewidth]{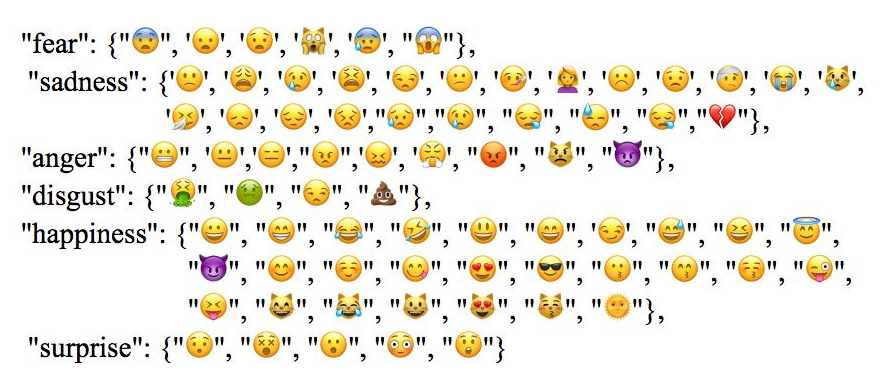}
\label{tabl}
\end{figure}
\label{pic1}

\begin{multicols}{2}
The classification of the texts’ emotions was carried out using the created dictionary. By counting the number of encounters of each emotion in the text, vectors of 7 emotions were generated for each sentence.

Text vectorization was carried out using our own NLP framework - BRAIN2NLP\footnote{Russian cognitive processor for automated labelling of natural text. The main functionality of BRAIN2NLP includes language detection, stem extraction, morphological and syntactic labelling of text, extraction of nominal entities and definition of a semantic vector of words.}. The text vectorization was based on the BERT language model trained by Google. Thus, at the output, pairs of the following form were obtained: 512-dimensional vector BERT to seven-dimensional vector of emotion. In total, more than 100,000 such pairs were constructed. In the process, it turned out that the resulting dataset contains a lot of conflicting data (collisions). Collisions are examples of data, where the same properties of an object lead to different classes or to a significant number of classes. The automation of the collision search process required the development of a small script. The algorithm of its work is given below.

\end{multicols}

\begin{algorithm}[H]
\SetAlgoLined
 \caption{Finding collisions in a dataset}
\begin{itemize}
\item[1.] 	Determine the allowed fraction of the k-number of classes to which objects in the same cluster can belong.
\item[2.]	Classify dataset objects by the vector of their properties (features).
\item[3.]	Assign the same ID to dataset objects belonging to the same cluster.
\item[4.]	Add up the values by class for objects with the same ID.
\item[5.]	Estimate for each ID the mathematical expectation of the vector of values by class.
\item[6.]	Calculate the number of Z values of the class vector that is greater than or equal to the expected value.
\item[7.]	Evaluate each cluster, and where $Z > k$ mark it as a collision.
\item[8.]	Assign a collision mark to all objects in the cluster.
\end{itemize}

\end{algorithm}

\begin{multicols}{2}
For the purpose of "human" marking, 10,000 texts were selected, among which there were only 6913 examples without collisions with a length of no more than two sentences. The experts were asked to assign no more than 2 emotions to each of these texts. But the dataset formed in this way also contained minor examples of collisions, which were also eliminated \cite{1}. As a result, a golden dataset with 2813 examples was obtained.

\section{Search for Architecture}

BERT-vectors are successfully used for classification tasks of texts \cite{9} and their fragments \cite{5}. Considering this fact, we have chosen these vectors as a basic description of the space of text properties, which can be used to predict the emotion vector.

During the search for the optimal solution for the classification of emotions, different variants of embedding length BERT (lengths 768 and 512) and different sizes of datasets were tested. Also, we have tested a certain number of negative samples, various dictionaries for the extraction of emoticons, normalization of the dataset, different model architectures, loss functions and activation functions. As a result of this work, the following hypothesis was formulated and confirmed in practice: \textit{It is possible to improve the quality of classification by abandoning the attempt to normalize the unbalanced initial data - the number of examples for a set of classes, replacing it with uniformly generated BERT vectors for each set of classes.}

Among all tested neural network architectures, Generative adversarial network (GAN) \cite{6} showed the best results - an architecture consisting of a generative network and a discriminative network configured to work against each other. Discriminative models learn the boundary between the classes; generative model generates new data instances. A collision free dataset (Dataset\_1) was used for training the generative network and for final testing of the discriminative network. In the formulation of the study, the vector of emotions has 7 dimensions, we denote it as V1. Table \ref{tab1} shows a list of combinations of vector V1 values found in Dataset\_1.

\end{multicols}

\begin{table}[h!]
\centering
\caption{Combinations of emotion vectors Dataset\_1}
\begin{tabular}{|c|c|c|c|c|c|c|c|}
\hline
\#          & \textbf{fear} & \textbf{sadness} & \textbf{anger} & \textbf{disgust} & \textbf{calm} & \textbf{happiness} & \textbf{surprise} \\ \hline
\textbf{0}  & 0             & 0                & 0              & 0                & 0             & 0                  & 1                 \\ \hline
\textbf{1}  & 0             & 0                & 0              & 0                & 0             & 1                  & 0                 \\ \hline
\textbf{2}  & 0             & 0                & 0              & 0                & 0             & 1                  & 1                 \\ \hline
\textbf{3}  & 0             & 0                & 0              & 0                & 1             & 0                  & 0                 \\ \hline
\textbf{4}  & 0             & 0                & 0              & 0                & 1             & 0                  & 1                 \\ \hline
\textbf{5}  & 0             & 0                & 0              & 0                & 1             & 1                  & 0                 \\ \hline
\textbf{6}  & 0             & 0                & 0              & 0                & 1             & 1                  & 1                 \\ \hline
\textbf{7}  & 0             & 0                & 0              & 1                & 0             & 0                  & 0                 \\ \hline
\textbf{8}  & 0             & 0                & 0              & 1                & 0             & 0                  & 1                 \\ \hline
\textbf{9}  & 0             & 0                & 0              & 1                & 1             & 0                  & 0                 \\ \hline
\textbf{10} & 0             & 0                & 1              & 0                & 0             & 0                  & 0                 \\ \hline
\textbf{11} & 0             & 0                & 1              & 0                & 0             & 0                  & 1                 \\ \hline
\textbf{12} & 0             & 0                & 1              & 0                & 0             & 1                  & 0                 \\ \hline
\textbf{13} & 0             & 0                & 1              & 0                & 0             & 1                  & 1                 \\ \hline
\textbf{14} & 0             & 0                & 1              & 0                & 1             & 0                  & 0                 \\ \hline
\textbf{15} & 0             & 0                & 1              & 0                & 1             & 0                  & 1                 \\ \hline
\textbf{16} & 0             & 0                & 1              & 0                & 1             & 1                  & 0                 \\ \hline
\textbf{17} & 0             & 0                & 1              & 1                & 0             & 0                  & 0                 \\ \hline
\textbf{18} & 0             & 0                & 1              & 1                & 0             & 0                  & 1                 \\ \hline
\textbf{19} & 0             & 1                & 0              & 0                & 0             & 0                  & 0                 \\ \hline
\textbf{20} & 0             & 1                & 0              & 0                & 0             & 0                  & 1                 \\ \hline
\textbf{21} & 0             & 1                & 0              & 0                & 0             & 1                  & 0                 \\ \hline
\textbf{22} & 0             & 1                & 0              & 0                & 1             & 0                  & 0                 \\ \hline
\textbf{23} & 0             & 1                & 0              & 0                & 1             & 0                  & 1                 \\ \hline
\textbf{24} & 0             & 1                & 0              & 0                & 1             & 1                  & 0                 \\ \hline
\textbf{25} & 0             & 1                & 0              & 1                & 0             & 0                  & 0                 \\ \hline
\textbf{26} & 0             & 1                & 1              & 0                & 0             & 0                  & 0                 \\ \hline
\textbf{27} & 0             & 1                & 1              & 0                & 0             & 0                  & 1                 \\ \hline
\textbf{28} & 0             & 1                & 1              & 1                & 0             & 0                  & 0                 \\ \hline
\textbf{29} & 1             & 0                & 0              & 0                & 0             & 0                  & 0                 \\ \hline
\textbf{30} & 1             & 0                & 0              & 0                & 0             & 0                  & 1                 \\ \hline
\textbf{31} & 1             & 0                & 0              & 0                & 0             & 1                  & 1                 \\ \hline
\textbf{32} & 1             & 0                & 0              & 0                & 1             & 0                  & 0                 \\ \hline
\textbf{33} & 1             & 0                & 0              & 0                & 1             & 0                  & 1                 \\ \hline
\textbf{34} & 1             & 0                & 0              & 1                & 0             & 0                  & 0                 \\ \hline
\textbf{35} & 1             & 0                & 1              & 0                & 0             & 0                  & 1                 \\ \hline
\textbf{36} & 1             & 1                & 0              & 0                & 0             & 0                  & 0                 \\ \hline
\textbf{37} & 1             & 1                & 1              & 0                & 0             & 0                  & 0                 \\ \hline
\end{tabular}
\label{tab1}
\end{table}

\begin{multicols}{2}
Since 37 combinations\footnote{Not entirely confident in "Les 36 situations dramatiques" concept, we decided to generate all possible combinations of emotions.} do not make up a complete list of 128 combinations, the missing ones had to be generated.

For this purpose, based on Dataset\_1, we have trained a generator model for known combinations of emotions. Then the combinations V1 were initiated, where the initial - has the values [0, 0, 0, 0, 0, 0, 0], and the final one is [1, 1, 1, 1, 1, 1, 1]. The total number of vectors V1 was 128. Using the generator, 128 BERT - V2 vectors were obtained corresponding to various V1 combinations.

\begin{figure}[h!!]
\centering
\includegraphics[width=0.7\linewidth]{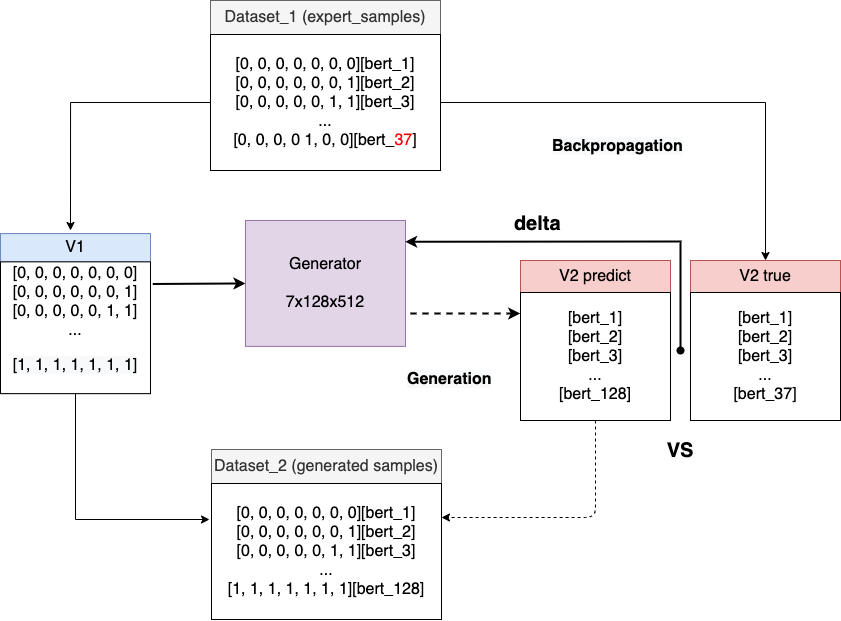}
\caption{The process of generator training with the formation of Dataset\_2}
\label{pic2}
\end{figure}

\begin{figure}[h!!]
\centering
\includegraphics[width=0.7\linewidth]{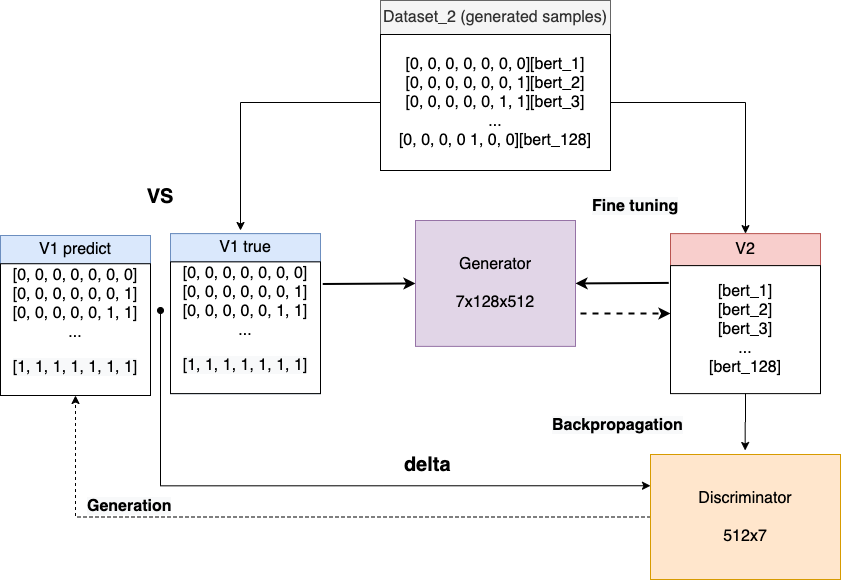}
\caption{The process of joint training of the Generator and the Discriminator}
\label{pic3}
\end{figure}

Thus, a new dataset (Dataset\_2) was formed, where each vector V1 corresponded to the generated vector V2 (Figure 1). Dataset\_2 was used for fine-tuning the Generator model and Discriminator training, predicting the emotion vector from the BERT vector of the text. Figure \ref{pic2} shows the process of sequential training of the Generator, and Figure \ref{pic3} shows the Discriminator.
\end{multicols}

\begin{multicols}{2}

\begin{figure}[h]
\begin{minipage}[h]{0.47\linewidth}
\center{\includegraphics[width=1\linewidth]{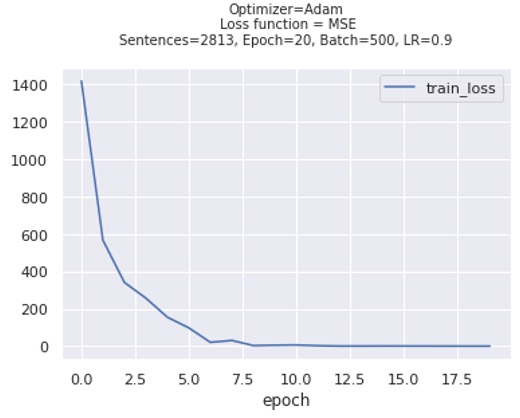}} \\a) 
\end{minipage} 
\hfill
\begin{minipage}[h]{0.47\linewidth}
\center{\includegraphics[width=1\linewidth]{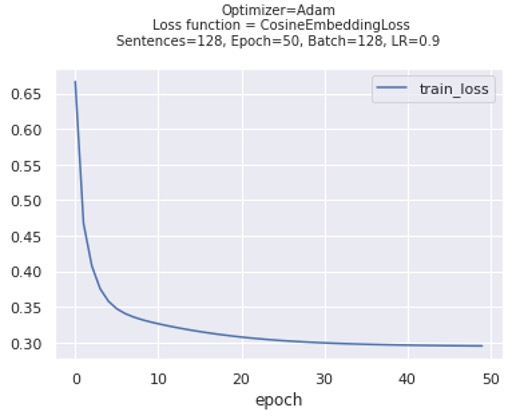}} \\b) 
\end{minipage}
\caption{Gradient descent of a) Generator model; b) Discriminator model}
\label{pic4}
\end{figure}

Figure \ref{pic4}  shows graphs of the "descent rate" when training the Generator and the Discriminator. The large value of the generator error is explained by the fact that a simple product of vectors is used, and the error is considered by MSE. The opposite situation is with the discriminator, where not only the output value is normalized, but also the error is calculated by the cosine measure. 
\end{multicols}

\begin{multicols}{2}

In order to move to a non-negative real number, the output vector of the discriminator value was normalized as follows 

\begin{equation}
M_{ij}^{\ast}=\dfrac{M_{ij}-min\vert M_{ij}\vert }{\sum_{M_{ij} \in V} ( M_{ij} -min\vert M_{ij}\vert  ) }
\end{equation}

where V is a vector of forecast values.

\section{Results}

Model training was performed on 70\% of the data with testing on the remaining 30\%. The evaluation of the quality of the model is illustrated in Figure 4. Comparative results of the model training process are presented in Table \ref{tab2}.
\end{multicols}

\begin{table}[]
\centering
 \quad \caption{Accuracy rating}
\begin{tabular}{|l|l|l}
\cline{1-2}
\textbf{Emotion} & \textbf{Accuracy} & \multirow{9}{*}{\begin{minipage}[h]{0.5\linewidth}
\center{\includegraphics[width=1\linewidth]{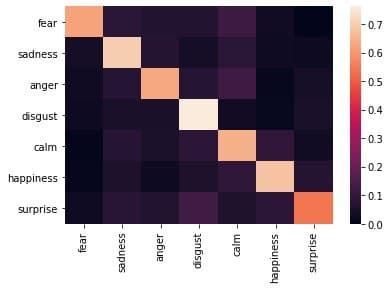}} \\ Figure 4: The matrix of the predictive distribution and the real value of the Discriminator based on Dataset\_1\end{minipage}} \\ \cline{1-2}
\textbf{fear} & 0.62 &  \\ \cline{1-2}
\textbf{sadness} & 0.70 &  \\ \cline{1-2}
\textbf{anger} & 0.63 &  \\ \cline{1-2}
\textbf{disgust} & 0.76 &  \\ \cline{1-2}
\textbf{calm} & 0.64 &  \\ \cline{1-2}
\textbf{happiness} & 0.68 &  \\ \cline{1-2}
\textbf{surprise} & 0.54 &  \\ \cline{1-2}
\multicolumn{1}{|r|}{\textbf{MEAN}} & 0.65 &  \\ \cline{1-2}
\end{tabular}
\label{tab2}
\vspace{5cm}
\end{table}

\begin{table}[h]
\caption{The result of the model training process}
\begin{tabular}{|c|c|c|c|c|c|c|c|c|c|c|c|c|}
\hline
\multirow{2}{*}{Network} & \multicolumn{3}{c|}{Dataset} & \multicolumn{2}{c|}{Architecture} & \multicolumn{7}{c|}{Learn} \\ \cline{2-13} 
 & \begin{tabular}[c]{@{}c@{}}\begin{sideways}Collisions\end{sideways}\end{tabular} &  \begin{sideways}Train \end{sideways}&\begin{sideways}Test\end{sideways}  & \begin{tabular}[c]{@{}c@{}}\begin{sideways}1-st layer \end{sideways}\end{tabular} & \begin{tabular}[c]{@{}c@{}}\begin{sideways}2-nd layer \end{sideways}\end{tabular} & \begin{tabular}[c]{@{}c@{}}\begin{sideways}Weight \end{sideways}\begin{sideways} Initialization\end{sideways}\end{tabular} & \begin{tabular}[c]{@{}c@{}}\begin{sideways}Activation \end{sideways}\begin{sideways} Function \end{sideways}\end{tabular} &\begin{sideways}Error\end{sideways}  & \begin{tabular}[c]{@{}c@{}}\begin{sideways} Gradient \end{sideways}\begin{sideways}descent \end{sideways}\end{tabular} & \begin{sideways}Epochs \end{sideways}& \begin{tabular}[c]{@{}c@{}}\begin{sideways} Train accuracy \end{sideways}\\ \end{tabular} & \begin{tabular}[c]{@{}c@{}}\begin{sideways}Test accuracy \end{sideways} \end{tabular} \\ \hline
\begin{tabular}[c]{@{}c@{}}Generator\end{tabular} & Yes & 4855 & 2082 & 7x128 & 128x512 & Random & No & MSE & ADAM & 10 & 0.99 & 0.99 \\ \hline
\begin{tabular}[c]{@{}c@{}}Discriminator\end{tabular} & Yes & 89 & 39 & 512x7 & No & \begin{tabular}[c]{@{}c@{}}Metod \\    FM*\end{tabular} & \begin{tabular}[c]{@{}c@{}}Cosine \\ similarity\end{tabular} & \begin{tabular}[c]{@{}c@{}}Cosine\\    \\ Loss\end{tabular} & ADAM & 50 & 0.75 & 0.81 \\ \hline
\begin{tabular}[c]{@{}c@{}}Generator\end{tabular} & No & 1969 & 844 & 7x128 & 128x512 & Random & No & MSE & ADAM & 10 & 0.96 & 0.9 \\ \hline
\begin{tabular}[c]{@{}c@{}}Discriminator\end{tabular} & No & 89 & 39 & 512x7 & No & \begin{tabular}[c]{@{}c@{}}Metod \\    FM*\end{tabular} & \begin{tabular}[c]{@{}c@{}}Cosinus \\ similarity\end{tabular} & \begin{tabular}[c]{@{}c@{}}Cosine\\    \\ Loss\end{tabular} & ADAM & 50 & 0.74 & 0.87 \\ \hline
\end{tabular}
\label{tab3}
*FM Method (frequency-matrix) is a method for forming a frequency matrix based on pairs of property and class vectors. The original dataset of the emotion vector and BERT vector pairs of text was divided into two matrices of emotion vectors and BERT vectors. Then, by multiplying the obtained matrices, a frequency matrix is formed.
\end{table}

\begin{multicols}{2}
Based on the input data, the model makes a forecast in the form of a 7-dimensional vector of emotions. To be able to compare it with gold dataset, it was converted into 2-dimensional vector by selecting emotion classes with maximum values. If at least one of the two emotions coincided with the gold labelling, then it was considered that the emotion was predicted correctly. The accuracy of the forecast for such an assessment varies depending on the class of emotion (see Table \ref{tab3}). The average accuracy of emotion predictions is 0.65.

The results of the model for different words with different emotional sentiment are shown in Table \ref{tab4}.

\end{multicols}

\begin{table}[h]
\centering
\caption{An example of model’s evaluation the emotional sentiment of a sentence}
\label{tab4}
\begin{tabular}{|l|l|l|l|l|l|l|l|l|l|}
\hline
\multicolumn{1}{|c|}{\multirow{2}{*}{\textbf{-}}} & \multicolumn{1}{c|}{\multirow{2}{*}{\textbf{ Example of sentence}}} & \multicolumn{1}{c|}{\multirow{2}{*}{\textbf{\begin{tabular}[c]{@{}c@{}}\\ Expert’s\\ Emotion\end{tabular}}}} & \multicolumn{7}{c|}{\textbf{Predicted   emotion by the Model}} \\ \cline{4-10} 
\multicolumn{1}{|c|}{} & \multicolumn{1}{c|}{} & \multicolumn{1}{c|}{} & \multicolumn{1}{c|}{\textbf{\begin{sideways} fear \end{sideways}}} & \multicolumn{1}{c|}{\textbf{\begin{sideways}sadness \end{sideways}}} & \multicolumn{1}{c|}{\textbf{\begin{sideways} anger \end{sideways}}} & \multicolumn{1}{c|}{\textbf{\begin{sideways} disgust \end{sideways}}} & \multicolumn{1}{c|}{\textbf{\begin{sideways} calm \end{sideways}}} & \multicolumn{1}{c|}{\textbf{\begin{sideways}  happiness\end{sideways}}} & \multicolumn{1}{c|}{\textbf{\begin{sideways}  surprise\end{sideways}}} \\ \hline
1 & \begin{tabular}[c]{@{}l@{}}There   is danger all around, \\ it's   scary to go outside.\end{tabular} & \begin{tabular}[c]{@{}l@{}}fear,\\ \\ sadness\end{tabular} & \textbf{0.49} & 0.12 & \textbf{0.17} & 0.13 & 0.00 & 0.00 & 0.09 \\ \hline
2 & \begin{tabular}[c]{@{}l@{}}It is   very sad that nothing \\ can   be done. The city \\ burned   down and everyone  \\ died\end{tabular} & \begin{tabular}[c]{@{}l@{}}sadness,\\ fear\end{tabular} & 0.18 & \textbf{0.27} & 0.21 & \textbf{0.27} & 0.00 & 0.00 & 0.07 \\ \hline
3 & \begin{tabular}[c]{@{}l@{}}I am   super angry, \\ I'm   in a fury! \\ You're   dead, bastards!\end{tabular} & anger & \textbf{0.21} & 0.15 & \textbf{0.22} & 0.19 & 0.00 & 0.12 & 0.11 \\ \hline
4 & \begin{tabular}[c]{@{}l@{}}The   soup was terrible, \\ I've   never tasted such  \\ disgusting   meal. I've been \\ sick   of it for an hour.\end{tabular} & \begin{tabular}[c]{@{}l@{}}disgust,\\ anger\end{tabular} & 0.17 & \textbf{0.26} & 0.15 & \textbf{0.30} & 0.00 & 0.07 & 0.06 \\ \hline
5 & \begin{tabular}[c]{@{}l@{}}Relax   and listen to nature, feel this \\ light   summer wind. The \\ silence   calms better than a \\ thousand   words.\end{tabular} & \begin{tabular}[c]{@{}l@{}}calm,\\ happiness\end{tabular} & 0.03 & \textbf{0.20} & \textbf{0.19} & 0.14 & 0.14 & 0.30 & 0.00 \\ \hline
6 & \begin{tabular}[c]{@{}l@{}}What could be better than \\ watching   the sunset together \\ with   close friends! \\ Just   wonderful!\end{tabular} & \begin{tabular}[c]{@{}l@{}}happiness,  \\ calm\end{tabular} & 0.01 & 0.05 & 0.10 & 0.03 & 0.00 & \textbf{0.63} & \textbf{0.18} \\ \hline
7 & \begin{tabular}[c]{@{}l@{}}John   won more than \\ \$1   million in the lottery, \\  that   was a pleasant surprise \\ for   him.\end{tabular} & \begin{tabular}[c]{@{}l@{}}surprise,\\ happiness\end{tabular} & 0.00 & 0.04 & 0.03 & 0.15 & 0.07 & \textbf{0.47} & \textbf{0.25} \\ \hline
\end{tabular}
\end{table}

\begin{multicols}{2}

Based on the study of the data on model’s performance, the following conclusions were made:
\begin{itemize}
\item[-]	in all cases, the model correctly identified at least one of the dominant emotions, a little less often – two of the two.
\item[-]	the model does not define the calm emotion well, in most cases its value is the lowest.
\item[-]	Intends to show more emotions than necessary (see examples in the first five sentences of Table \ref{tab4}).
\end{itemize}

That is, the model "gives a chance" to emotions that were not defined in the gold dataset. We consider this to be a special property of the model rather than a bug, since the model was trained on a larger variety (of generated data) than those observed by experts. This is a very important consequence that needs to be taken into account when reconstructing complete data based on expert assessments.

\section{Conclusion}

Among all the architectures that were tested, GAN performed better than the others (see Table \ref{tab5}). When compared with the models of emotions of some other authors, our model showed the best result. The F1-measure value is 0.025 higher than the "goemotions-pytorch" model, and the accuracy value is 0.0053 higher than the "Emotion Detection and Recognition from Text Using Deep Learning" model, but 0.19 lower than the "Bidirectional LSTM with a Convolutional Neural Network"model. In the latter case, it is worth saying that in our model the number of defined emotions is seven against four in "Bidirectional LSTM with a Convolutional Neural Network".

\end{multicols}

\begin{table}[h]
\centering
\caption{Comparing the quality of different emotion models}
\label{tab5}
\begin{tabular}{|l|c|c|c|}
\hline
\multicolumn{1}{|c|}{\textbf{Model}} & \textbf{Emotions} & \textbf{F1 macro} & \textbf{Accuracy} \\ \hline
\begin{tabular}[c]{@{}l@{}}«Bidirectional LSTM with a Convolutional Neural \\ Network» by Rodrigo Masaru Ohashi (2019)\end{tabular} & 4 & - & 0.84 \\ \hline
\begin{tabular}[c]{@{}l@{}}«Emotion Detection and Recognition from Text Using\\ Deep Learning» by Chew-Yean Yam (2015)\end{tabular} & 5 & - & 0.64 \\ \hline
«GoEmotions-pytorch» by Jangwon Park (2020) & 7 & 0.61 & - \\ \hline
\textbf{GAN BERT-EMO model (2020)} & \textbf{7} & \textbf{0.64} & \textbf{0.65} \\ \hline
\end{tabular}
\end{table}
\begin{multicols}{2}

The hypothesis about the possibility of improving the quality of classification by refusing to normalize the class distribution of a real sample and replacing it with a sample obtained by generating all possible combinations of classes (BERT vectors) for a finite number of variables was fully justified. The authors have not encountered such an approach in other publications. At the same time, it opens up new perspectives for machine learning with minimizing human involvement by automatically generating "big data" for training neural networks on "small" incomplete data.  
\end{multicols}

\bibliographystyle{unsrt}  

\end{document}